\documentclass[letterpaper, 10 pt, conference]{ieeeconf}  % Comment this line out
\IEEEoverridecommandlockouts
\usepackage[utf8]{inputenc}
\usepackage{authblk}
\usepackage{graphicx}
\usepackage{fixltx2e}
\graphicspath{ {./Pix/} }
\title{\LARGE \bf
DRDr: Automatic Masking of Exudates and Microaneurysms Caused By Diabetic Retinopathy Using Mask R-CNN and Transfer Learning
}

\author[1]{Farzan Shenavarmasouleh}
\author[1]{Hamid R. Arabnia}
\affil[1]{Department of Computer Science, University of Georgia, Athens, Georgia, United States \authorcr {\{fs04199, hra\}@uga.edu}\vspace{1.5ex}}

\begin{document}
\maketitle
\thispagestyle{empty}
\pagestyle{empty}

%%%%%%%%%%%%%%%%%%%%%%%%%%%%%%%%%%
\begin{abstract}
This paper addresses the problem of identifying two main types of lesions - Exudates and Microaneurysms - caused by Diabetic Retinopathy (DR) in the eyes of diabetic patients. We make use of Convolutional Neural Networks (CNNs) and Transfer Learning to locate and generate high-quality segmentation mask for each instance of the lesion that can be found in the patients' fundus images. We create our normalized database out of e-ophtha EX and e-ophtha MA and tweak Mask R-CNN to detect small lesions. Moreover, we employ data augmentation and the pre-trained weights of ResNet101 to compensate for our small dataset. Our model achieves promising test mAP of 0.45, altogether showing that it can aid clinicians and ophthalmologist in the process of detecting and treating the infamous DR.

\vspace{\baselineskip}

\end{abstract}
%%%%%%%%%%%%%%%%%%%%%%%%%%%%%%%%%
\section{\textbf{INTRODUCTION}}
Diabetic retinopathy is a major cause of vision impairment, and eventually vision loss in the world; especially among working-aged individuals. Its diagnosis can be done by analyzing color fundus images by experienced clinicians to identify its presence and the significance of the damages that it has caused. Fundus images are the results of screenings. The procedure is easy and it can easily and safely be done via retinal photography in every clinic with the proper tools. If detected soon enough, diabetic retinopathy (DR) can be treated via laser surgeries. But, the demand is increasing much more rapidly than the supply. The annual number of patients is growing and each patient requires frequent screenings. Each of these images needs to be carefully analyzed by doctors. The task is innately time-consuming since the deficiencies are usually extremely small and require careful examination and the doctors need to find and weight countless features for each image. The thing is from all the patients being screened annually, only 25.2\% have diabetic retinopathy and are referred to ophthalmologist \cite{massin2008ophdiat} and it begs the question as to whether it is possible to use the time and resources more sufficiently?

Luckily, the answer to the above question is yes. During the past few years, Machine Learning and notably Deep Learning have shown high potential in helping health care and they can be used to aid doctors in detecting and predicting the development of various illnesses. In fact, machine learning and deep learning approaches have already helped numerous researchers to overcome some of the difficulties in the aforementioned task at hand. However, the majority of the previous work in this part lies in the machine learning section, and classification and especially binary classification was the primary subject of interest. In other words, previous work was mostly devoted to finding a way to automatically identify whether a patient has diabetic retinopathy or not.

Decenci{\`e}re et al. \cite{decenciere2013teleophta} leveraged a big dataset extracted from OPHDIAT \cite{massin2008ophdiat}, a teleophthalmology network, during 2008-2009 and the help of three experts to tackle this issue. They merged features extracted from images with the patients' contextual data such as age, weight, diabetic type, and the number of years of DR and altogether could predict whether the patient needs to be referred or not. Bhatia et al. \cite{bhatia2016diagnosis} and Antal et al. \cite{antal2014ensemble} used ensembles of machine learning techniques, namely Decision Trees, Support Vector Machines (SVM), Adaboost, Naïve Bayes, and Random Forests, on Messidor dataset. Usher et al. \cite{usher2004automated} and Gardner et al. \cite{gardner1996automatic} employed Neural Networks to perform the task of classification for them. The former, utilized candidate lesions, their position, and their type as the inputs of the Neural Network and the latter used Neural Networks and pixel intensity values.

In Priya et al. \cite{priya2013diagnosis}, the authors explored Probabilistic Neural Networks (PNN) along with Naïve Bayes and SVM. They employed Adaptive Histogram Equalization, Discrete Wavelet Transform, Matched Filter Response, Fuzzy C-Means Segmentation, and Morphological Processing on top of the Green channel of the images for their preprocessing phase. The train/test split was questionable though, as out of 350 total images, the authors used 250 of them for test and only 100 for the training.

Sopharak et al. \cite{sopharak2010machine} made use of Naïve Bayes, SVM and K Nearest Neighbors (KNN) classifiers to detect exudates in pixel level. This task was traditionally being dealt with by using region growing and thresholding \cite{liu1997automatic, ege2000screening, sinthanayothin1999automated}. They selected 15 handpicked features and operated on them. However, their dataset was extremely small with only 39 images.

Several attempts were made to extend the level of classification as well and drag the level of severity to this task too. Lachure et al. \cite{lachure2015diabetic} utilized SVM and KNN to classify images of Messidor and DB-reet into 3 classes. Their model could tell whether a fundus image is normal or not; and if abnormal, whether it is grade 1 or 3. Roychowdhury et al. \cite{roychowdhury2013dream} used Gaussian Mixture Model (GMM), KNN, SVM, Adaboost along with feature selection to classify images of Messidor in 4 classes. Acharya et al. \cite{acharya2008application} and Adarsh et al. \cite{adarsh2013multiclass} also used SVM to deal with this problem and classified patients into 5 classes.

All of the forenamed approaches required an external feature extraction phase. Authors needed to manually perform multiple morphological operations, apply various filters to the images, and use techniques such as region growing and thresholding to extract features one by one and then fuse them with other contextual data, if any, and then use the resulting files as inputs for the different classifiers. 

With the advancement of deep learning and specifically Convolutional Neural Networks (CNN), network architectures solely designed to enhance working with images, the task of feature extraction turned into an implicit phase instead. Gargeya et al. \cite{gargeya2017automated} made use of CNN to identify healthy and unhealthy patients using a huge dataset with 75 thousand images. Gulshan et al. \cite{gulshan2016development} employed Inception V3, a more complex type of CNN, to classify images of EyePACS-1 and Messidor-2 into two categories. And finally, Pratt et al. \cite{pratt2016convolutional} harnessed CNN and a huge publicly available Kaggle dataset to classify the fundus images into 5 classes.

In this paper, we approach the problem from another angle and address the problem of automatically identifying the deficiencies caused by diabetic retinopathy in fundus images together with their exact shape and location. We modify and leverage a CNN-based model that can identify and use the intricate features in the available images to detect, locate, and most importantly mask and label two important lesion types, namely microaneurysms and exudates. Due to the automatic behavior of our approach, it can easily fit into clinical systems and aid clinicians in the process of identifying unhealthy patients while saving them plenty of time. It has the potential to both be incorporated into retinal cameras and/or be used as a post-photography tool for optometrists and ophthalmologists.

The remainder of the paper is organized as follows: First, we touch base with the related works that are aligned with our interests. Then, we fully explain our methodology, including how we handle our limited available data effectively. Next, we show the experiments that we have performed and illustrate our results. Finally, we conclude with the discussion and future work.

%%%%%%%%%%%%%%%%%%%%%%%%%%%%
\section{\textbf{Related Work}}
\subsection{\textbf{Convolutional Neural Networks}}
Computer Vision is the branch of computer science which its ultimate goal is to imitate the behavior and functionality of human eyes. It assists computers to analyze images and videos and ultimately understand objects that are present in them. Thanks to the advancement of Deep Learning in the past few years, we are now able to handle this task very well. Since AlexNet \cite{krizhevsky2012imagenet} won the ImageNet image classification competition in 2012, Convolutional Neural Networks (CNN) have become the go-to approach for any task that required dealing with images. In fact, nowadays, CNNs are so powerful that they even surpass humans' performance on the ImageNet challenge. Image captioning \cite{amirian2019image}, Learning from Observation \cite{soanssa}, and Embodied Question Answering \cite{das2018embodied} ,to name a few, are some of the very high-level tasks that implicitly use CNNs as one of their core components. Besides, researchers in the field of Meta-Learning and Neural Architecture Search (NAS) are constantly trying to find an innovative way to further improve the performance of these systems \cite{mohammadi2019parameter, xie2019exploring}.

Image classification, Object Localization, Object Detection, Semantic Segmentation, and Instance Segmentation are 5 main Computer Vision problems; sorted by their level of difficulty in ascending order. In Image Classification problem, usually exists an image with a single main object in it and the goal is to predict what category that image belongs to. A tad more challenging task is Object Localization. In object localization, the image usually contains one or more objects from the same category and the model's goal is to output the location of those objects as bounding boxes, a rectangular box around the object, besides predicting the category that they all are affiliated with. 

As impressive as they look like, these tasks are not remotely as complex as what humans visual understanding and eyes are capable of doing.

Next is Object detection and recently a huge breakthrough has happened in it. CNNs work similarly to human eyes and they are able to detect edges and consequently define boundaries of the objects. Hence, they could be used to detect objects of different kinds in a given image. However, to do so, it is required to apply them to a massive number of locations with varieties of scales on each image, making it extremely time-consuming. As a result, an extensive amount of research has been done to tackle this issue.

R-CNN (Region-based CNN) \cite{girshick2014rich} solves the aforementioned issue by making use of a region proposal module. This module proposes a collection of candidate bounding boxes, also known as Regions of Interests (ROIs), using the Selective Search technique. The pixels corresponding to each of these boxes are then fed into a pre-trained modified version of AlexNet to check if any object is present inside that box. On the very last layer of this CNN lies an SVM that judges whether the pixels represent an object or not and if yes, what is the category that goes with them. At last, if an object is found in the box, the box is tightened to best fit the object dimensions.

Training an R-CNN model is hard and time-consuming because approximately 2,000 ROIs are proposed for every single image and all of them need to be fed to the CNN individually. Besides, three different networks are ought to be trained separately.

Fast R-CNN \cite{girshick2015fast} solved both of these problems. Normally, many of the regions that are proposed for further examination overlap with each other and hence this causes the CNN phase to do so many redundant computations. Fast R-CNN overcomes this issue by using only one CNN  per image to compute all the features at once. The result is then shared and used by all the 2000 proposals, reducing the computational time significantly. This technique is called Region of Interest Pooling (RoIPool) in the original paper.

To tackle the second issue, Fast R-CNN united all the three models into one single network to enable jointly training of all of them. The SVM classifier was exchanged with a Softmax layer to handle the task of classification and a regression layer was added in parallel to that to find and yield the best bounding box for each object.

However, Fast R-CNN still used the Region Proposal method using selective search to find the ROIs, which turned out to be the bottleneck of the overall process.

Faster R-CNN \cite{ren2015faster} was proposed to resolve this issue. The authors' main intention was to replace the selective search phase with something more efficient. They argued that the image feature maps that were already calculated with the forward pass of the CNN could be fed directly to a Fully Convolutional Network (FCN) on top of them to perform the task of region proposal instead of running a separate selective search algorithm. This newly added FCN was called Region Proposal Network (RPN) in the paper and this way, ROIs could be proposed almost for free, fixing the last problem present in the system.

As wonderful as object detection is, it still could not understand and provide us the actual shape of the objects and stops at delivering the bounding box only. This task, however, is where Image Segmentation comes into play and tackles the issue by creating a pixel-wise mask for each object. The task of image segmentation itself can be done in two main ways.
In Semantic Segmentation, every pixel in the image needs to be assigned to a predefined class. In addition, all the pixels corresponding to a class are treated the same and are given an identical color and thus the differences among different object instances that belong to one class are disregarded. By contrast, in Instance Segmentation, each instance of the same class is treated discretely and given a unique color and label.

Mask R-CNN \cite{he2017mask} is a model developed for the task of image instance segmentation. It extends Faster R-CNN, goes one step further, and specializes in generating pixel-level masks for each object in excess of finding the bounding box and the class label. It, creatively, adds another FCN on top of RPN, altogether creating a new parallel branch to the Fast R-CNN model which outputs a binary mask for the object found in a given region. It is also worth noting that the authors needed to slightly modify the RoIPool to fix the problem of location misalignment caused by its quantization behavior. They called the modified technique RoIAlign.

\subsection{\textbf{Transfer Learning}}
Humans are really good at transferring their knowledge across tasks. We rarely learn a certain task from scratch and instead, we tend to leverage our previous knowledge that we have acquired in the past in some similar activity or topic. By doing so, we utilize and accelerate our new learning process. Traditional machine learning and deep learning algorithms are designed to work in insulation and learn to handle only a domain-specific task. To make a model work for another task or domain, the entire model has to be retrained from scratch, and thus not taking advantage of the previously learned task will result in consuming so much more time and resources than required, as plenty of redundant operations are needed to take place. Besides, often a huge amount of labeled data are required to learn a specific task in a supervised manner. Preparing such datasets are innately hard as it takes a long time to collect and then label them manually. And sometimes constructing them are nearly impossible for some domains. 

Transfer learning is the proposed solution to overcome this issue and facilitate the knowledge sharing process among different tasks. It suggests reusing parts of the model which have been already trained on a similar task as the foundation for the new task at hand. This approach has proved to be extremely helpful in cases where no good or large enough dataset is available for the target domain, but a fairly good one exists for the source domain. In addition, it saves a lot of time and computational power as the pre-trained weights are employed and the model only needs to learn the last few layers and barely fine-tune the other ones if necessary.

%%%%%%%%%%%%%%%%%%%%%%%%%%
\section{\textbf{Methodology}} % Dataset and Experiment Setup

% A publicly available Kaggle dataset (For Severity and Later work and since the dataset was huge and the images were collected from all over the world the zoom and the lightening were very different.)

\subsection{\textbf{Dataset}}
Diabetic retinopathy causes different types of deficiencies in the patients' eyes such as exudates, hemorrhages, aneurysms, cotton wool spots, and abnormal growth of blood vessels to name a few. There are many publicly available datasets that could be found online, some small and some really huge. Often, big datasets are used for learning complex tasks to avoid overfitting and ensure the reproducibility of the research result \cite{farzan2019}. However, we needed a dataset that could offer masks for the type of defect which it corresponds to. We came across the e-ophtha \cite{decenciere2013teleophta} which had two separate masked datasets; one for exudates and one for aneurysms; and both were manually annotated by ophthalmology experts. E-ophtha EX contains 47 images with exudates and 35 images with no lesion, while e-ophtha MA provides 148 images with microaneurysms or small hemorrhages and 233 images with no lesion. 

We only made use of the images with lesions present in them, which altogether summed up to a total of 195 fundus pictures, all having another black and white image as their mask. We shuffled and splitted them in to train, validation, and test sets with 155, 20, 20 images in each of them respectively.

\begin{figure*}[ht]
    \centering
    \includegraphics[width=16cm]{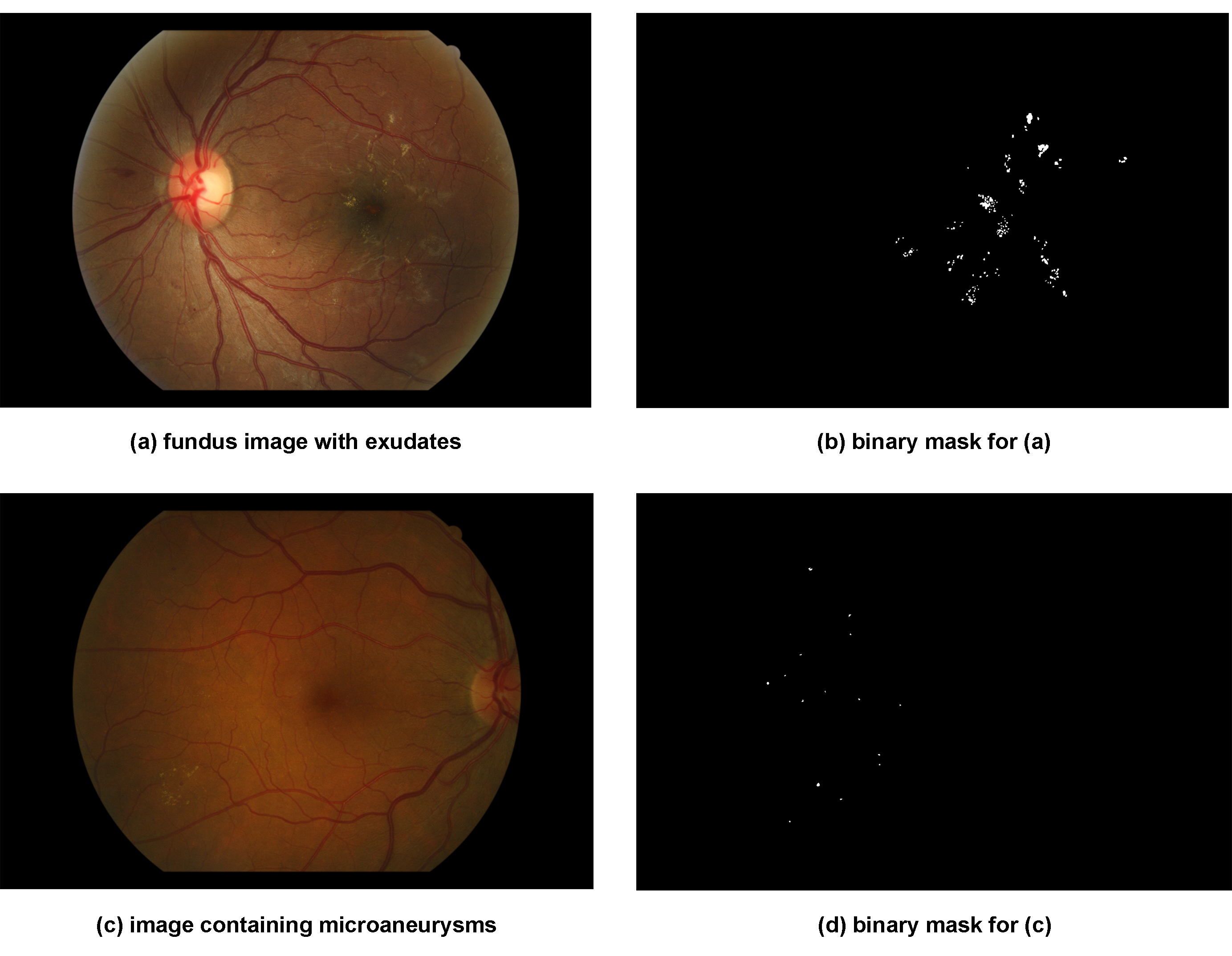}
    \caption{(a) an image from e-ophtha EX containing exudates. (b) binary mask showing the position of exudates in image (a). (c) an image from e-ophtha MA containing microaneurysms. (d) binary mask showing the location of microaneurysms in picture (c).}
    \label{fig:Fig1}
\end{figure*}

\subsection{\textbf{Preprocessing}}
The images in the datasets were collected from different clinics with different fundus photography facilities and this had resulted in having varying lighting and pixel intensity values in them, collectively creating unimportant differences among pictures that would have been misleading for the CNN model if left unaltered. Hence, a preprocessing phase was required to counterbalance this issue. 

First, we employed OpenCV \cite{opencv_library} to crop the images and trim the extra blank space from them. Next, to make the dataset even more homogeneous, we morphed the eyes into perfect circles and removed the extra margins once more. The colors needed to be normalized and enhanced as well, so we performed a weighted sum, and for each image, we applied Gaussian blur (sigma X = 20) on it and added it to its original version. We assigned weights of 4 and -4 to the original and blurred images respectively. The gamma was also set to 128. Finally, the images were resized to 1024x1024 pixels.

All of the above operations were concurrently applied to the masks as well, to preserve the exact scale and position that they signify in the image. But, the masks, especially the ones which corresponded to microaneurysms were extremely small and only consisted of a handful of pixels (between 1 to 5) as opposed to the total image dimension of 1024x1024. This would have been a tremendously hard task for the model to learn. To tackle this issue, the masks were dilated 2 times with a kernel size of 5x5 to make them big enough to be identifiable by the network.

Also, all the instances of a certain defect were shown in one single mask in the dataset. To make them usable, we needed to create a separate binary mask for each instance present in the image. Thus, first, we found all the contours in a given mask. Then, we detached the instances and constructed an exclusive binary mask for each of them. Finally, class ids were assigned to the masks to indicate which defect each of them represents.

\begin{figure*}[ht]
    \centering
    \includegraphics[width=17.7cm]{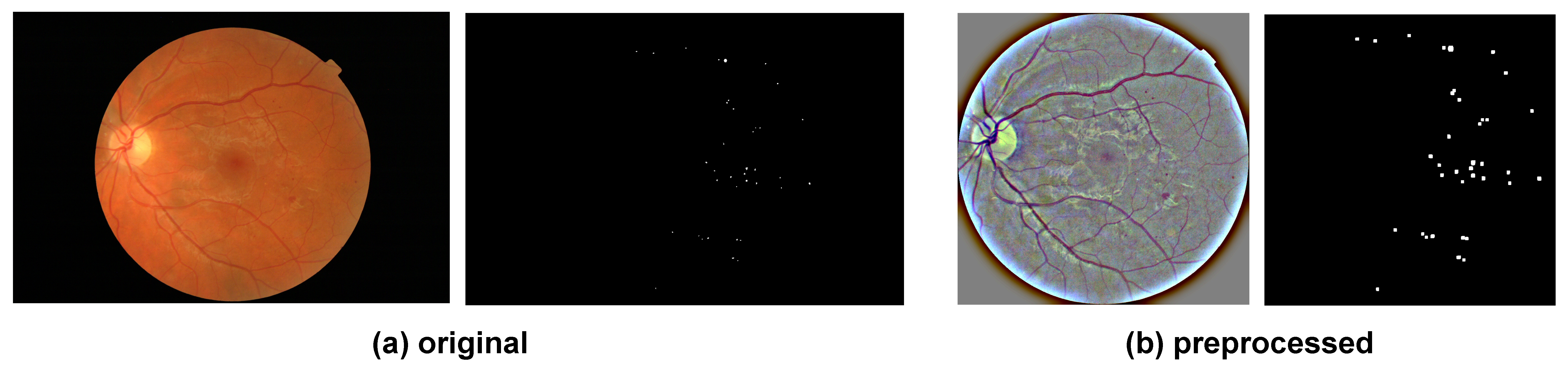}
    \caption{(a) an example of the original image from e-ophtha MA and its mask. (b) resulting image and its mask after the preprocessing phase.}
    \label{fig:Fig2}
\end{figure*}

\subsection{\textbf{Training and Implementation Details}}
We started with the original implementation of Mask R-CNN for Keras which was made publicly available by its authors. By default, the model's hyperparameters were configured to find the medium to large objects in the image. However, as mentioned before, even after dilating the masks, they only consisted of a few pixels and were really small relative to the complete picture. Hence, we had to make alterations to the model to make it suitable for our task at hand. RPN anchor sizes had to be decreased to enable the model to find deficits as small as 8 pixels in size. Since lesions were small and could be found anywhere in the image, the number of anchors to be trained were increased from 256 to 512, the number of ROIs per image was raised to 512, and the maximum number of final detections was set to 256. Also, we needed to reduce the minimum confidence and threshold required for the model to accept a detection. We disabled the mini-mask feature to avoid any mask resize as we had enough memory and didn't have to sacrifice accuracy for the memory load. Also, we defined the number of classes to be three; one for the background, and two more for exudates and microaneurysms. Adam optimizer was found to be more effective than the default stochastic gradient descent as well. So, it was employed instead to help the model converge to the optimal point faster.

We found out that Mask R-CNN can easily overfit the training set if used naively. As for our first way out, We made use of data augmentation. It comprised random vertical and horizontal flips, 90 degrees clockwise and counterclockwise rotations, and translations and scalings along x and y axes; all of which been applied to the input training images on the fly with the help of the CPU, in parallel to the main training which was being done on our Nvidia 2080Ti GPU to accelerate the process even more.

Our dataset was small, and thus, because of the aforementioned reasons, the best way to counteract this was to employ transfer learning. We put the pre-trained weights of ResNet101 \cite{he2016deep} which were originally been trained on Microsoft COCO dataset \cite{lin2014microsoft} into service. The model was then trained for 65 epochs with the learning rates of 0.0001, 0.00001, and 0.000001 for two 25, and one 15 epochs respectively to fine-tune all the initial weights and make them suitable for our new task. The process in total took about 15 hours to finish.

%%%%%%%%%%%%%%%%%%%%%%%%%%
\section{\textbf{Experiments and Results}}
It is conventional to evaluate and measure the performance of segmentation models with IoU and mAP. Intersection over Union (IoU) calculates the area of the overlap that happens between the predicted bounding box and the actual mask and then divides it to the area of the union of those two. A completely correct bounding box will result in IoU of 1. A threshold is also set to accept IoUs above it as correct predictions. The percentage of correct predictions out of all predicted bounding boxes is called precision. Recall, on the other hand, is the percentage of correct predictions out of all objects present in the image. As more and more predictions are made the precision will decrease due to false positives, but the recall will increase. These two are calculated for different thresholds and then averaged to find out the AP (average precision) for a given image. The mean of APs across all the images in the dataset is referred to as mAP (mean average precision). 

To calculate the mAP for our model, we first needed to fix an issue. Our model was deliberately trained on both tasks simultaneously; giving it the ability to find both types of lesions in a given fundus image at the same time. However, the masks that we had from the datasets were only associated with one type of lesion and for the most part, there was no overlap between the two datasets. If used this way, it would have caused our model to get a lower mAP. The reason behind it was that in the test phase, given a fundus image, our model would have predicted and masked both types of lesions, but the mask that it was being compared to was only showing one type, altogether making the evaluation agent think that the lesions from the other type are all false positives and hence reduce the precision.

To fix this, when the predictions were made for a given image by our model, we passed them through a filter to only keep the ones that are associated with the type that is marked in the corresponding mask image. This enabled us to test our model in a fair way and see how accurate the exudates and microaneurysms are being predicted individually. 

Due to the innate complexity of the task and the extremely small size of the lesions that had to be found, we had previously decreased our model's minimum prediction confidence hyperparameter to 35. Hence, we used it along with two more standard thresholds that are usually used in the task of instance segmentation. As a result, 35, 50, and 75 were chosen as our three thresholds to calculate the results with. Results of the evaluation can be found in Table \ref{tab:res-table}.

\begin{table}[h]
\centering
\resizebox{7cm}{!}{%
\begin{tabular}{|c|ccc|}
\hline
           & mAP\textsubscript{35} & mAP\textsubscript{50} & mAP\textsubscript{75} \\ \hline
Train      & 0.5408 & 0.5217 & 0.3032 \\ 
Validation & 0.5113 & 0.4780 & 0.2563 \\ 
Test       & 0.4562 & 0.4370 & 0.2071 \\ \hline
\end{tabular}%
}
\caption{mAP for train, validation, and test sets created from e-ophtha EX and e-ophtha MA datasets}
\label{tab:res-table}
\end{table}

To the best of our knowledge, our work is the first to employ instance segmentation models to identify and mask the lesions in the eye and help to diagnose the infamous diabetic retinopathy. Given the complexity of the task, our model performed extremely well and we call our result a success.

\begin{figure*}[ht]
    \centering
    \includegraphics[width=17.7cm]{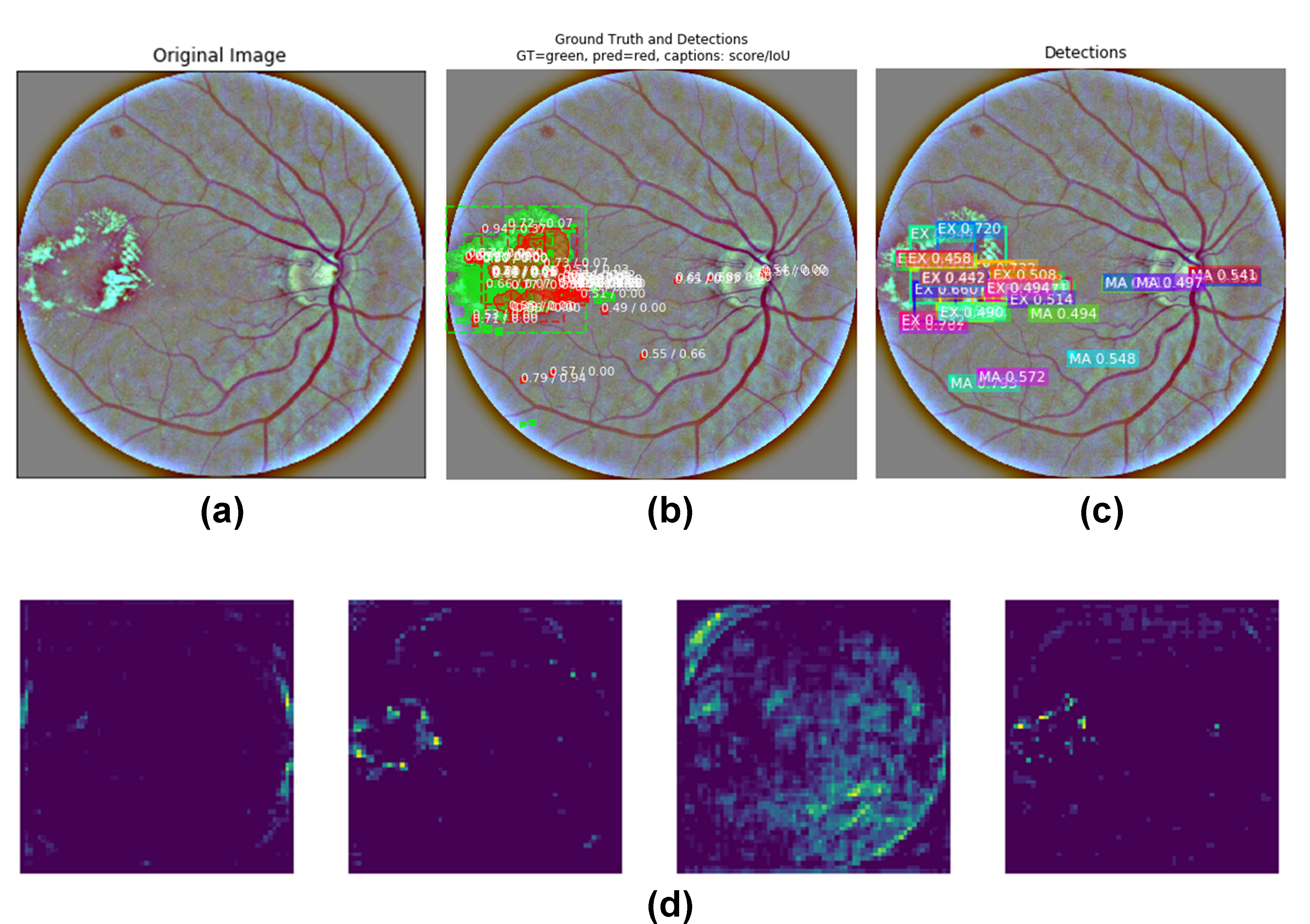}
    \caption{(a) original image (b) predicted masks, bounding boxes, their score and IoU (c) Types of the lesions detected and their scores. (d) sample activations of a few layers of the model}
    \label{fig:Fig3}
\end{figure*}

%%%%%%%%%%%%%%%%%%%%%%%
\section{\textbf{Conclusion and Future Work}}
We have presented a simple, yet efficient approach to detect, locate, and generate segmentation masks for exudates and microaneurysms which are two types of lesions that diabetic retinopathy causes in eyes. Unlike most of the previous work, our model is capable of automatically extracting useful features related to the scale of our work, and learn to perform the task in an end-to-end manner. Moreover, due to its fast predictions, it has the potential to be easily incorporated into health care facilities. 

Future work should consider using our model to identify the severity of DR in patients' eyes, exploring other instance segmentation architectures and their ensembles, and creating a better and bigger dataset with more types of lesions.

% Result of this segmentation on different datasets.
% It can be further improved if it’s annotated by experts, more images are used, and the hyperparameters get fine-tuned even more.
% UNET + ensemble + ... maybe
% Maybe use the areas, number of each type of deficiencies and use them to detect the severity using a neural network. -> In classification they report sensitivity which is the percentage of correctly identified patients having Diabetic Retinopathy. Accuracy, which is …, and finally specificity which is correctly saying that a patient is healthy.

%%%%%%%%%%%

\section{\textbf{Conflict of Interest}}
The authors declare that there is no conflict of interest regarding the publication of this article.

%%%%%%%%%%%%%%%%%%%%%%%%%%%%%%%%%%%%%%%%%%%%%%%%%%%%%%%%%%%%%%%%%%%%%%%%%%%%%%%%

\bibliography{bib} 
\bibliographystyle{ieeetr}

\end{document}